\documentclass[11pt]{article}

\usepackage[preprint]{arxiv}
\usepackage[T1]{fontenc}
\usepackage[utf8]{inputenc}
\usepackage{lmodern}
\usepackage{microtype}
\usepackage{amsmath}
\usepackage{amssymb}
\usepackage{booktabs}
\usepackage{graphicx}
\usepackage{array}
\usepackage{tabularx}
\usepackage{ltablex}
\usepackage{enumitem}
\usepackage{caption}
\usepackage{xurl}
\usepackage[hidelinks]{hyperref}
\usepackage[numbers,sort&compress]{natbib}

\graphicspath{{figures/}}

\newcommand{\repositoryurl}{\href{https://github.com/AyoubJadouli/Quantbot-Research-Framework}{companion repository}}
\keepXColumns

\hypersetup{
  pdftitle={Predictive Extrema, Unprofitable Policies: An AI-Assisted Audit of Candle-Based Binance Spot Timing Models},
  pdfauthor={Ayoub Jadouli},
  pdfsubject={Cryptocurrency forecasting, chronological evaluation, artifact provenance, AI-assisted research},
  pdfkeywords={cryptocurrency forecasting, extrema detection, transaction costs, chronological evaluation, backtest overfitting, AI-assisted research}
}

\title{Predictive Extrema, Unprofitable Policies:\\
An AI-Assisted Audit of Candle-Based Binance Spot Timing Models}
\author{
  Ayoub Jadouli\\
  Computer Science and Smart Systems, Faculty of Sciences and Technology\\
  Abdelmalek Essa\^adi University, Tangier, Morocco\\
  \texttt{ayoubjadouli@gmail.com}
}
\date{July 20, 2026}

\begin{document}
\maketitle

\begin{abstract}
We audit whether candle-based machine-learning models can turn predictions of cryptocurrency extrema or short-horizon outcomes into positive Binance Spot paper policies after assumed costs.  Numerical results come from scripted fixed-seed model runs and deterministic simulators; human-supervised AI agents supported the July 20 evidence-integrity revision through literature retrieval, separately tasked critique, artifact reconciliation, documentation, and source packaging, not trading decisions.  The strongest later-period evidence, conditional on extensive predecessor search, is negative: an unchanged ten-pair mandatory-daily selector lost 6.72\% over 19 July cycles at an assumed 31-bps completed-cycle cost, with 3 wins and 16 losses.  In short model-specific July evaluations, the validation-selected local-minimum policy returned -1.79\%, while the local-maximum sell-to-cash/re-entry policy underperformed continuous holding by 2.80\%; their gross mean advantages of 11.11 and 12.21 bps were below even the 21-bps stress.  A Gurgul-inspired, OHLCV-only daily adaptation attained minimum/maximum ROC AUC of 0.874/0.896 but average precision of only 0.134/0.116 and lost 44.30\% over seven cycles, versus -41.20\% for buy-and-hold.  A forensic audit also downgraded an earlier One4All ``30-day holdout'': its dates had influenced prior architecture work, its four-hour outcome horizon was not purged at split boundaries, it used same-close entry, and its raw result directories were absent.  Across the tested, mostly exploratory protocols, event-ranking performance did not establish positive executable policy value.  Every operational decision remains \texttt{NO\_TRADE}.
\end{abstract}


\section{Introduction}

Financial prediction and trading-policy evaluation are different problems.  A classifier may rank rare future extrema well while a stateful policy still buys too late, exits poorly, trades too often, or loses more on its misses than it gains on its hits.  Transaction costs, next-executable-price timing, abstention, path ordering, portfolio state, and adaptive model selection determine whether a prediction has economic value.  This distinction is especially important for volatile crypto-assets and rare-event labels.

This paper reports a negative study of candle-based Binance Spot timing models.  The research program tested cross-pair selectors, shared recurrent and convolutional models, after-minimum and first-touch targets, forced daily selection, fourteen-model local-extrema campaigns, an OHLCV-only adaptation of a published paired-extrema policy, and slower rotation controls.  Several older families supply historical context but are not quantitative ledger rows because their protocols differ and some source bundles are missing.  The aim is not to identify the best-looking historical rule.  It is to determine what survives a conservative evidence audit after accounting for chronology, costs, support, search budget, and artifact provenance.

Four questions organize the revision:
\begin{enumerate}[leftmargin=*]
  \item Can high discrimination on rare extrema translate into positive policy return?
  \item Does an unchanged intraday policy retain value on later observations?
  \item How do adaptive search, consumed periods, execution conventions, and missing artifacts change the strength of a claim?
  \item What can AI agents usefully automate in a research workflow without becoming market-decision agents?
\end{enumerate}

The paper makes three contributions.  First, it provides an artifact-backed prediction--policy comparison in which strong ROC AUC coexists with negative net policy outcomes.  Second, it adds a disjoint twelve-day extension to a frozen July selector, bringing the unchanged prospective campaign to nineteen cycles without retuning.  Third, it documents a self-audit that removed silent numeric fallbacks and downgraded an overstated holdout.  That correction is part of the result: negative evidence is useful only when its provenance and limitations are reported honestly.

All claims are exchange-, universe-, period-, and simulator-specific.  This is not evidence that machine learning cannot support crypto trading in other settings, and it is not investment advice.  It is evidence that the tested policies do not justify deployment.

\section{Related Work}

\subsection{Cryptocurrency forecasting and trading after costs}

Cryptocurrency machine-learning studies have reported predictive or trading gains under several horizons and market conditions \citep{alessandretti2018anticipating,sebastiao2021forecasting}.  A recent closely related arXiv preprint evaluates hourly BTC-USDT with XGBoost, LSTM, and iTransformer over 27 walk-forward folds \citep{bysik2026costs}.  Its naive sign policies fail at 10 bps, while a cost-aware magnitude filter restores profitability in selected configurations.  That contrast is important: the present negative result should not be generalized beyond its five-minute, multi-pair, label, execution, and cost assumptions.  It instead agrees with the narrower conclusion that forecast-to-action translation and turnover control can matter more than architecture alone.  A crypto-specific deep-reinforcement-learning arXiv preprint likewise treats apparent backtest profitability as a selection-risk problem \citep{gort2023drl}.

The daily paired-extrema campaign is inspired by Gurgul, Lessmann, and H\"ardle \citep{gurgul2025extrema}.  Their study combines financial, blockchain, GitHub, search, macroeconomic, and text/social variables with separate local-minimum and local-maximum forecasts.  The local campaign preserves the target/policy idea but not the multimodal data or exact model implementations.  It is therefore an OHLCV-only, cost-aware adaptation, not a replication or reproduction of their full study.

\subsection{Prediction, decisions, and abstention}

Predictive loss is not generally aligned with downstream decision loss.  Predict-then-optimize research formalizes this mismatch and motivates objectives tied to the eventual action \citep{elmachtoub2022smart}.  Selective classification similarly studies the coverage--risk trade-off when a model may abstain \citep{geifman2017selective}.  These ideas are directly relevant here: sparse acceptance sometimes looked less negative, while mandatory coverage admitted weak trades.  Because extrema are rare, average precision and prevalence are reported beside ROC AUC; precision--recall summaries are more informative than ROC curves alone under severe imbalance \citep{saito2015precision}.

\subsection{Chronology, adaptive search, and reproducibility}

Out-of-sample evaluation is standard for serial and potentially non-stationary data.  Standard cross-validation can be justified only under additional conditions, including suitable error dependence assumptions \citep{bergmeir2018cv}; empirical comparisons of time-series evaluation schemes further caution that protocol choice changes error estimates \citep{cerqueira2020evaluating}.  Financial data snooping, reality checks, deflated performance, and formal backtesting protocols all warn that repeated strategy search inflates apparent performance \citep{lo1990datasnooping,sullivan1999datasnooping,white2000reality,hansen2005spa,bailey2014backtest,bailey2014deflatedsharpe,harvey2016crosssection,arnott2019backtesting}.  This paper does not claim a formal family-wise correction.  It instead reports search budgets, separates evidence classes, and refuses promotion.

Finally, AI-assisted research must be distinguished from agentic trading.  A recent arXiv survey of LLM trading-agent studies found sparse reporting of time-consistent splits, explicit costs, and reproducible artifacts \citep{xia2026agentic}.  The agents described here operate around manuscript maintenance and evidence checking; they do not perceive a live market or emit tradable actions.

\section{Evidence Scope and Safety Boundary}

\subsection{Common boundary, family-specific simulators}

Every strategy is research-only.  Models emit scores or paper proposals; family-specific deterministic simulators apply portfolio and exit rules.  No LLM, strategy, manuscript script, or result can construct or transmit a Binance mainnet order.  Unknown state fails closed to \texttt{NO\_TRADE}.  The common element across campaigns is this safety boundary, not a single execution simulator.

Numerical results originate in stored code and artifacts.  Agents helped locate, challenge, and communicate that evidence, while the human author remained responsible for scope, interpretation, and release.  Section~\ref{sec:ai-assistance} discloses that workflow; it is research-process documentation rather than an empirical contribution or a source of trading authority.

\subsection{Two-dimensional evidence taxonomy}

Evidence strength has two separate dimensions:
\begin{itemize}[leftmargin=*]
  \item \textbf{Artifact support:} raw summary/trade/model records, narrative-only records, or missing inputs.
  \item \textbf{Data status:} frozen prospective, model-specific later-period, exposed/consumed diagnostic, or descriptive control.
\end{itemize}
An artifact-backed result is not automatically pristine, and a chronologically later interval is not untouched if it influenced earlier architecture or policy choices.  Table~\ref{tab:provenance} records the local audit status.  Missing empirical inputs are never replaced with fallback numbers.

\begin{table}[htbp]
  \centering
  \caption{Artifact provenance after the July 20 fail-closed audit.}
  \label{tab:provenance}
  \resizebox{\textwidth}{!}{\begin{tabular}{llll}
\toprule
Evidence family & Expected record & Local status & Manuscript use \\
\midrule
July mandatory selector & summary, model, trades & present & Quantitative evidence \\
Local minimum / maximum & summaries, models, trades, manifests & present & Quantitative diagnostics \\
Daily paired adaptation & summary, models, events, predictions & present & Quantitative diagnostic \\
Slow rotation v3 & summary, decisions, monthly returns, hashes & present & Consumed descriptive control \\
One4All 30-day after-min & summary and trade directory & absent & Narrative archival record only \\
Earlier rotation plot inputs & three equity CSV paths & absent & Removed from manuscript \\
\bottomrule
\end{tabular}
}
\end{table}

\begin{figure}[htbp]
  \centering
  \includegraphics[width=\textwidth]{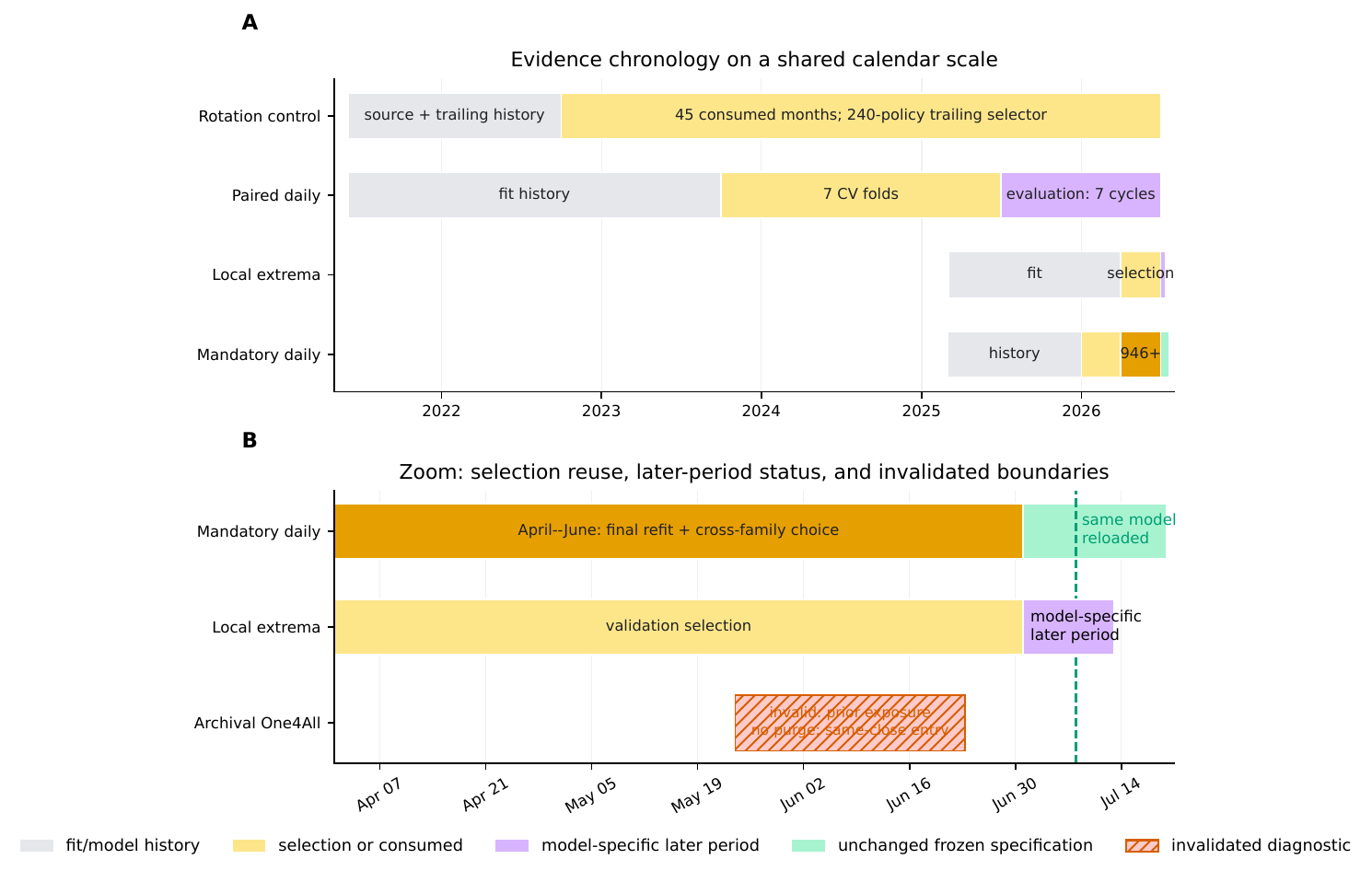}
  \caption{Chronology and data-status audit, drawn on a common calendar scale.  Panel A maps the artifact-backed mandatory-daily, local-extrema, paired-daily, and rotation campaigns; colored spans distinguish fitting/history, adaptive selection or consumed data, model-specific later evaluation, and the unchanged July specification.  Panel B enlarges April--July 2026, including the dual use of April--June for final cross-family choice among at least 946 enumerated candidates and final refitting through June 30, the 19-cycle July freeze, the 12-day model-reload extension, and the archival One4All interval.  The hatched One4All span is an invalidated diagnostic---not holdout evidence---because of prior architecture exposure, an unpurged four-hour label horizon, same-close entry, and absent raw result directories.  The figure reports experimental contracts and evidence status, not policy returns.}
  \label{fig:chronology}
\end{figure}

Figure~\ref{fig:chronology} puts the four artifact-backed campaigns on one calendar and then enlarges the 2026 boundaries that would otherwise be visually negligible.  The distinction between fit, selection, consumed historical evidence, model-specific later periods, and an unchanged frozen specification is part of the experimental result: chronological order alone does not make a period independent of prior research decisions.

\subsection{Audited experimental contracts}

Table~\ref{tab:contracts} supplies the operational detail behind the chronology: universe and bar size, target construction, support, exact fit/selection/evaluation roles, execution timing, cost stress, search budget, benchmark, and final evidence status.  The rows deliberately retain family-specific return semantics instead of implying that all five campaigns share one simulator.  The archival One4All row is included to make its exclusion auditable, not to restore it to the quantitative result set.

\begin{center}
  \captionsetup{hypcap=false}
  \captionof{table}{Experimental contracts and evidence status used in the July 20 audit.  Cost labels are round-trip assumptions: the intraday campaigns deduct them per completed cycle, paired daily applies half at each executed side multiplicatively, and rotation applies half per unit of one-way turnover.  Support counts refer to policy cycles or label-valid observations as identified in each row; heterogeneous returns should not be ranked across rows.}
  \label{tab:contracts}
\end{center}
\begingroup
\renewcommand{\theHtable}{contract-internal-\arabic{table}}
\begingroup
\scriptsize
\setlength{\tabcolsep}{3.5pt}
\renewcommand{\arraystretch}{1.15}
\begin{tabularx}{\textwidth}{@{}>{\raggedright\arraybackslash}p{0.14\textwidth}
  >{\raggedright\arraybackslash}X
  >{\raggedright\arraybackslash}X
  >{\raggedright\arraybackslash}X@{}}
\toprule
Campaign & Data, target, and support & Fit $\rightarrow$ selection $\rightarrow$ evaluation & Execution, search, comparison, and evidence status \\
\midrule
\textbf{Mandatory daily selector}
& \textbf{Universe/bar:} ten USDT Spot pairs; one-minute OHLCV.  \textbf{Target:} four-hour realized policy outcome with $+100/-150$ gross-bps target/stop or timeout; one ExtraTrees realized-net regressor ranks pairs.  \textbf{Support:} 19 mandatory July cycles (3 wins, 16 losses).
& \textbf{Fit:} 2025-03-01--2026-06-30.  \textbf{Select:} within-family on 2026-01-01--03-31; April--June was then reused for cross-family freezing after at least 946 enumerated candidates in eight campaign summaries.  \textbf{Evaluate:} 2026-07-01--19; the stored model was reloaded unchanged for July 8--19.
& \textbf{Policy/cost:} decide after 16:00 UTC completes, enter 16:01 next open, exactly one pair, at most four hours, adverse stop-first same-bar tie; 20/31/51-bps completed-cycle costs (31 primary).  \textbf{Benchmark:} cash/no-trade; no independent asset benchmark in the frozen summary.  \textbf{Status:} artifact-backed prospective failure conditional on extensive predecessor search; \texttt{NO\_TRADE}. \\
\midrule
\textbf{Local extrema: minimum / maximum}
& \textbf{Universe/bar:} BTC, ETH, SOL against USDT; five-minute OHLCV; 48-bar causal lookback.  \textbf{Target:} centered low/high with $b\in\{6,12,24\}$ bars on each side and at least a 40-bps rebound/pullback; $b=24$ selected.  \textbf{Support:} 10,293 held-out rows per direction (146/165 positives); selected policies completed 9/15 cycles.
& \textbf{Fit:} 2025-03-03--2026-03-31.  \textbf{Select:} 2026-04-01--06-30, with label-horizon purging.  \textbf{Evaluate:} 2026-07-01--12.  July is short and model-specific; July 1--7 had already appeared in unrelated repository research.
& \textbf{Policy/cost:} next-five-minute-open action, $+60/-50$ gross-bps barriers, 24-bar horizon, stop-first tie; 21/31/51 bps (31 primary).  Minimum buys from cash; maximum sells an existing Spot holding and later repurchases, so its return is advantage versus continuous hold, not short P\&L.  \textbf{Search:} 14 models, three windows, and up to 14 threshold quantiles---238 validation policy comparisons per direction.  \textbf{Benchmark/status:} cash for minimum and continuous hold for maximum; every gate failed, least-loss diagnostic only; \texttt{NO\_TRADE}. \\
\midrule
\textbf{Paired daily extrema (Gurgul-inspired)}
& \textbf{Universe/bar:} BTCUSDT and ETHUSDT daily OHLCV, 2021-06-01 through the 2026-07-01 terminal open.  \textbf{Target:} separate next-day-open minimum/maximum labels over symmetric 7/14/21-day windows.  \textbf{Support:} 704 label-valid held-out symbol-days; seven completed policy cycles (14 transactions).
& \textbf{Select:} seven expanding three-month folds starting 2023-10-01 through 2025-04-01, each with a preceding 90-day threshold-calibration slice and purged training history.  \textbf{Final fit/calibrate:} fit strictly before 2025-04-01; common threshold calibrated on 2025-04-01--06-30.  \textbf{Evaluate:} 2025-07-01--2026-06-30 once.
& \textbf{Policy/cost:} two 50\% sleeves start in cash; signals execute next daily open, conflicts preserve state, no short/borrowing, terminal liquidation; 21/31/51 bps (31 primary).  \textbf{Search:} 15 family/window profiles across seven folds and seven-threshold grids; 14-day HGB proxy selected by absolute return.  \textbf{Benchmark:} cost-matched equal-weight BTC/ETH buy-and-hold.  \textbf{Status:} artifact-backed OHLCV-only adaptation, not multimodal replication; holdout exposed by unrelated research; \texttt{NO\_TRADE}. \\
\midrule
\textbf{Consumed monthly rotation control}
& \textbf{Universe/bar:} ten USDT Spot pairs; daily OHLCV from 2021-06-01 through the 2026-07-01 terminal open.  \textbf{Policy signal:} completed-close, 30/45/60/90/120-day momentum and market/asset gates.  \textbf{Support:} 45 scored months (2022-10--2026-06), 15 target-active and 21 exposure-or-turnover months.
& No fixed fit/evaluation split: each scored month uses only the preceding 12 completed monthly windows, then executes the selected target at the next month open.  All 45 scored months and all observations through June 2026 are consumed historical research evidence; July 1 is liquidation only.
& \textbf{Execution/cost:} long-only, fully funded, drift-aware monthly targets, next-open rebalance and terminal liquidation; policy selection assumes 20 bps, then unchanged targets replay at 21/31/51 bps (31 primary), charging half-cost on one-way turnover.  \textbf{Search:} 240 past-only policies plus cash.  \textbf{Benchmarks:} fixed-60/top-one, equal-weight hold, and no-trade.  \textbf{Status:} positive but descriptive and concentration-sensitive (one month supplies 67.48\% of baseline net log gain); no prospective lock; \texttt{NO\_TRADE}. \\
\midrule
\textbf{Archival One4All diagnostic (invalidated)}
& \textbf{Universe/bar:} ten locally selected USDT pairs; five-minute candles; 96-bar input.  \textbf{Target/model:} four-hour after-minimum/outcome heads (plus a direct sell-30 variant) in one shared three-layer LSTM with residual feed-forward blocks.  \textbf{Support:} nominal 30-day interval, but auditable support is unavailable because the cited summary/trade directories are absent.
& \textbf{Nominal fit:} 2025-02-21 18:15--2026-02-22 00:15 UTC.  \textbf{Calibration:} 2026-02-22 00:15--05-24 07:45.  \textbf{Nominal evaluation:} 2026-05-24 07:45--06-23 07:45.  These dates had influenced prior architecture work, and the four-hour label horizon was not purged at either split boundary.
& \textbf{Execution/cost:} decide from a completed candle and fill at that same close; 20-bps completed-cycle deduction.  \textbf{Search:} 3,996 rank/top-$N$/stop/target profiles.  \textbf{Benchmark:} an equal-weight-hold comparison was reported only in narrative records.  \textbf{Status:} boundary-overlapped, optimistic-timing, missing-artifact archival diagnostic; excluded from quantitative results and never promotion evidence; \texttt{NO\_TRADE}. \\
\bottomrule
\end{tabularx}
\endgroup

\addtocounter{table}{-1}
\endgroup

\section{Data, Labels, and Simulation Assumptions}

\subsection{Data and universes}

All campaigns use public Binance Spot klines.  A kline contains open, high, low, close, volume, timing, and activity fields but not the within-bar transaction path \citep{binance2026klines}.  The main artifact-backed intraday campaigns use:
\begin{itemize}[leftmargin=*]
  \item three-pair five-minute local-extrema data (BTCUSDT, ETHUSDT, SOLUSDT), March 2025--July 12, 2026;
  \item ten-pair one-minute mandatory-daily data (ADA, AVAX, BNB, BTC, DOGE, ETH, SHIB, SOL, TRX, and XRP against USDT), March 2025--July 19, 2026;
  \item BTCUSDT and ETHUSDT daily candles for the paired-extrema adaptation, June 2021 through the terminal open on July 1, 2026, with outcomes scored through June;
  \item ten-pair daily candles for the consumed rotation control, June 2021 through the same July 1, 2026 terminal open.
\end{itemize}
The universes are not mutually controlled.  Longer-history availability and mechanical volume screening may introduce survivorship or selection bias, especially in rotation results.

\subsection{Costs and execution semantics}

Costs are assumptions, not measured realized commissions.  Legacy One4All diagnostics deduct 20 bps once per completed cycle.  The artifact-backed July local-extrema and mandatory-daily policies use a flat all-in 31-bps primary deduction per completed cycle.  The paired daily study treats 31 bps as a round trip and applies 15.5 bps multiplicatively at each executed side, including terminal liquidation; its buy-and-hold benchmark is cost-matched.  The rotation control likewise treats 31 bps as a round trip, but charges half-cost per unit of one-way rebalance turnover rather than a flat cycle deduction.  Local extrema and paired daily use 21/51-bps lower/upper stresses; rotation replays unchanged targets under the same round-trip labels; the prospective selector uses 20/51 bps.  Binance's published regular-user Spot schedule lists 0.100\% per maker or taker side, while actual commission depends on account, symbol, side, order type, and discounts \citep{binance2026fees,binance2026commission}.  The 21-bps floor roughly represents two 10-bps sides plus a 1-bps buffer; 31 and 51 bps add 10 and 30 bps of stress.  These are not measured decompositions of fee, spread, slippage, latency, impact, queue position, partial fills, or missed fills.

Artifact-backed July intraday policies decide from completed bars and enter at the next one-minute or five-minute open.  The daily adaptation also executes at the next daily open.  Stops and targets are evaluated from OHLC bars.  Across-bar ordering is chronological.  When target and stop are both reachable within one bar, their true order is unknowable from OHLCV; the simulators therefore use an adverse stop-first tie rule rather than assigning a favorable synthetic path.

For an executable entry price $p_e$, gross target $g$, gross stop $s$, completed-cycle cost $c$, and timeout $H$, the net target and stop outcomes are $(g-c)$ and $-(s+c)$ bps.  A timeout closes at the horizon price $p_H$:
\[
r_{\mathrm{timeout}}=10{,}000\left(\frac{p_H}{p_e}-1\right)-c.
\]
Each reported return states whether it is compounded portfolio return, cash-cycle advantage versus continuous holding, or a benchmark comparison.

\section{Experimental Campaigns}

\subsection{Frozen mandatory-daily prospective campaign}

The strongest later-period test uses a candidate whose within-family configuration was selected on January--March validation.  Before the July freeze, however, at least 946 enumerated candidates across eight July 9 mandatory-action campaign summaries---plus earlier searches---had been compared on April--June outcomes; the final candidate was chosen as the least-negative across those families.  After that choice, the frozen specification was refit through June 30, so April--June served both as consumed cross-family selection data and as part of the final fit.  The July result is prospective only conditional on that extensive adaptive history.  One ExtraTrees realized-net regressor ranks the ten-pair universe after the candle timestamped 16:00 UTC has completed, enters at the 16:01 open, holds at most four hours, and uses a +100-bps gross target and -150-bps gross stop.  The policy must select exactly one pair.  July 1--7 was evaluated once; on July 20, the exact stored model (SHA-256 \texttt{484d501b...f2eae}) evaluated the disjoint July 8--19 segment.  No universe, feature, target, stop, time, threshold, model, training cutoff, or cost changed.

\subsection{Local-minimum and local-maximum campaigns}

The three-pair campaigns use 48 completed five-minute bars of causal OHLCV, activity, indicator, price-location, and clock features.  For a radius $b\in\{6,12,24\}$ bars (30, 60, or 120 minutes on \emph{each} side), bar $t$ is a positive minimum/maximum only if its low/high is the extremum over the centered $2b+1$ span and the next $b$ bars contain at least a 40-bps rebound/pullback.  A paper action enters at the next open, scans target and stop for $b$ bars, and otherwise times out at the following open.  A validation-only scout freezes $b$.  Fourteen tabular and sequence variants then use that window: logistic regression, weighted linear SVM, random forest, histogram gradient boosting (HGB), Gaussian Bayes, MLP, CNN, GRU, LSTM, Transformer, CNN--GRU, attention CNN--LSTM, a price-location-attention hybrid, and GRU-feature boosting.

Training ends March 31, validation covers April--June, and the short model-specific evaluation covers July 1--12.  Label horizons are purged at boundaries.  Each model scans thresholds at up to 14 validation-score quantiles, with at least 30 validation cycles required for an eligible choice.  No window or model passed the predeclared positive-return/mean-return/drawdown gate; the attention CNN--LSTM is a least-loss diagnostic, not a validated candidate.  For minima, the policy enters next-open with +60/-50-bps gross target/stop.  For maxima, an already-held Spot position is sold next-open and later repurchased; reported return is cash-cycle advantage versus continuous holding, not short P\&L.  Each direction evaluates 238 validation policy combinations before the July slice.  The short July slices are model-specific/consumed rather than project-wide pristine; the minimum artifact specifically notes July 1--7 exposure.

\subsection{Gurgul-inspired daily adaptation}

From completed day $t$, the adaptation predicts separately whether the open of day $t+1$ is a BTC or ETH minimum/maximum over a symmetric $2w+1$-open span, where $w\in\{7,14,21\}$ days.  Fifteen family/window profiles are compared over seven expanding quarterly folds.  Absolute strategy return selected the 14-day HGB proxy even though its median cross-validation excess versus buy-and-hold was -18.60 percentage points.  A final April--June 2025 calibration freezes a common 0.80 score threshold; the model-specific July 2025--June 2026 period is then scored once.  Label horizons mature $w$ days after the target and are purged, signals execute next daily open, two equal sleeves start in cash, and the policy neither shorts nor borrows.  At the primary round-trip assumption, each executed buy or sell applies a multiplicative 15.5-bps side cost; terminal liquidation and the equal-weight BTC/ETH benchmark use the same convention.  Public OHLCV replaces the source paper's multimodal variables, and HGB/TFT-inspired models are proxies rather than exact XGBoost/TFT implementations.

\subsection{Consumed slow-rotation control}

The slow control selects among 240 past-only monthly policies using twelve trailing calibration months and next-open execution.  Its audited sample contains 45 scored months and primary 31-bps cost.  All dates are consumed historical evidence.  The control is included to show that a positive historical path can still fail a promotion audit because of sparse activity, concentration, universe bias, and absence of a prospective lock.

\subsection{Archival One4All diagnostic}

An earlier shared ten-pair LSTM used 96 five-minute bars, three LSTM layers, residual feed-forward blocks, and entry, target-touch, expected-net, tail-risk, market, and pair heads.  Its final sweep searched 3,996 rank/top-(N)/stop/target combinations.  Experience records report negative calibration-to-later-period translation, but this revision does not treat its values as confirmatory quantitative evidence.  Section~\ref{sec:forensic} explains why.

\section{Results}

\subsection{Artifact-backed evidence ledger}

Table~\ref{tab:evidence} reports heterogeneous results without pretending they are a controlled horse race.  Period, cost, support, return definition, and data status are explicit.

\begin{table}[htbp]
  \centering
  \caption{Artifact-backed results.  Metrics differ by policy and are not directly rank-comparable.}
  \label{tab:evidence}
  \resizebox{\textwidth}{!}{\begin{tabular}{llllll}
\toprule
Study & Period & Cost & Support & Result & Data status \\
\midrule
Frozen mandatory daily selector & 2026-07-01--19 & 31 bps & 19 cycles (3/16) & -6.72\% compounded & July continuation after cross-family April--June selection \\
Local-minimum selected model & 2026-07-01--12 & 31 bps & 9 cycles (5/4) & -1.79\% compounded & short model-specific holdout; partly exposed elsewhere \\
Local-maximum selected model & 2026-07-01--12 & 31 bps & 15 cycles (8/7) & -2.80\% cash-cycle advantage & short model-specific holdout; exposed elsewhere \\
Daily paired-extrema adaptation & 2025-07--2026-06 & 31 bps & 7 cycles (1/6) & -44.30\% vs -41.20\% buy-and-hold & model-specific holdout; exposed by unrelated research \\
Slow rotation baseline & 2022-10--2026-06 & 31 bps & 21 traded / 15 target-active months & 313.01\% compounded & consumed descriptive evidence; concentration-sensitive \\
\bottomrule
\end{tabular}
}
\end{table}

The unchanged mandatory selector is the clearest later-period result, conditional on the predecessor search disclosed above.  At 31 bps, July 1--19 compounded 100 USDT to 93.2815 USDT: -671.85 bps over 19 cycles.  Only 3 cycles were positive and mean net was -36.40 bps/cycle.  Conditional on independent Bernoulli trials, the descriptive Clopper--Pearson 95\% interval is 3.4--39.6\%; consecutive market-linked cycles need not satisfy that assumption.  The disjoint July 8--19 extension alone lost -354.28 bps with 3 wins and 9 losses.  The reporting cutoff and sample size were not preregistered, so this interim sample cannot estimate stable performance; it nevertheless supplies no positive later-period evidence.

\begin{figure}[htbp]
  \centering
  \includegraphics[width=\textwidth]{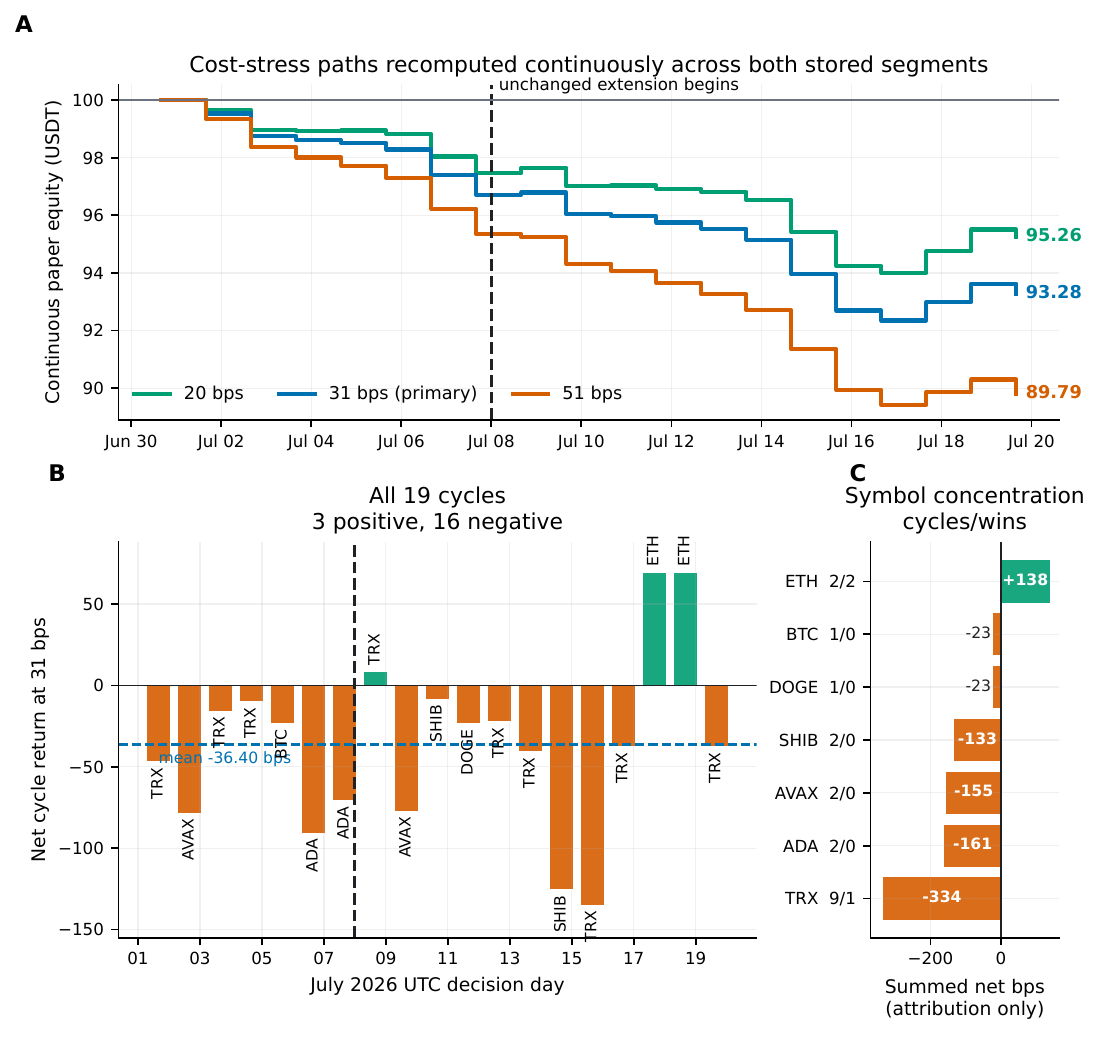}
  \caption{Unchanged mandatory-daily selector on all 19 July 2026 cycles.  Panel A recomputes continuous paper equity from the two stored trade segments under 20-, 31-, and 51-bps completed-cycle costs; the vertical divider marks the July 8 start of the disjoint extension, which reloads the same stored model without retuning.  Panel B plots every 31-bps cycle, including its selected pair and the -36.40-bps sample mean (3 positive, 16 negative).  Panel C sums 31-bps outcomes by selected symbol and reports each symbol's cycles and wins; these sums are an attribution, not separately investable or compounded portfolios.  The policy must choose one pair per day, and the 19-cycle path is frozen-specification evidence only conditional on extensive April--June cross-family selection; the cutoff and sample size were not preregistered.}
  \label{fig:prospective-july}
\end{figure}

Figure~\ref{fig:prospective-july} exposes every selected day instead of reducing the campaign to one endpoint.  At the primary cost, the daily outcomes ranged from -134.75 to +69.00 bps.  TRXUSDT was selected in 9 of 19 cycles and contributed -334.28 summed net bps; the attribution is descriptive and non-compounded, but it makes the policy's selection concentration visible.  The three cost paths remain negative, ending at 95.2579, 93.2815, and 89.7873 USDT under 20, 31, and 51 bps, respectively.

\clearpage

\subsection{Prediction--policy disconnect}

Table~\ref{tab:disconnect} aligns event discrimination with each model's paper-policy result.  The two post-hoc detector rows are diagnostic and cannot replace the validation-selected models.

\begin{table}[htbp]
  \centering
  \caption{Rare-event ranking versus economic policy value at 31 bps.  AP is average precision and Prev. is event prevalence.}
  \label{tab:disconnect}
  \resizebox{\textwidth}{!}{\begin{tabular}{llrrrrrr}
\toprule
Target / model & Status & AUC & AP & Prev. & Cycles & Mean net & Return \\
\midrule
Local minimum / attention CNN-LSTM & validation-selected & 0.729 & 0.043 & 1.42\% & 9 & -19.9 bps & -1.79\% \\
Local minimum / logistic & post-hoc detector diagnostic & 0.973 & 0.360 & 1.42\% & 15 & -27.2 bps & -4.02\% \\
Local maximum / attention CNN-LSTM & validation-selected & 0.750 & 0.053 & 1.60\% & 15 & -18.8 bps & -2.80\% \\
Local maximum / HGB & post-hoc detector diagnostic & 0.969 & 0.288 & 1.60\% & 45 & -39.9 bps & -16.50\% \\
Daily paired HGB (mean of min/max) & CV-selected adaptation & 0.885 & 0.125 & 2.56\% & 7 & -1460.6 bps & -44.30\% \\
\bottomrule
\end{tabular}
}
\end{table}

\begin{figure}[htbp]
  \centering
  \includegraphics[width=\textwidth]{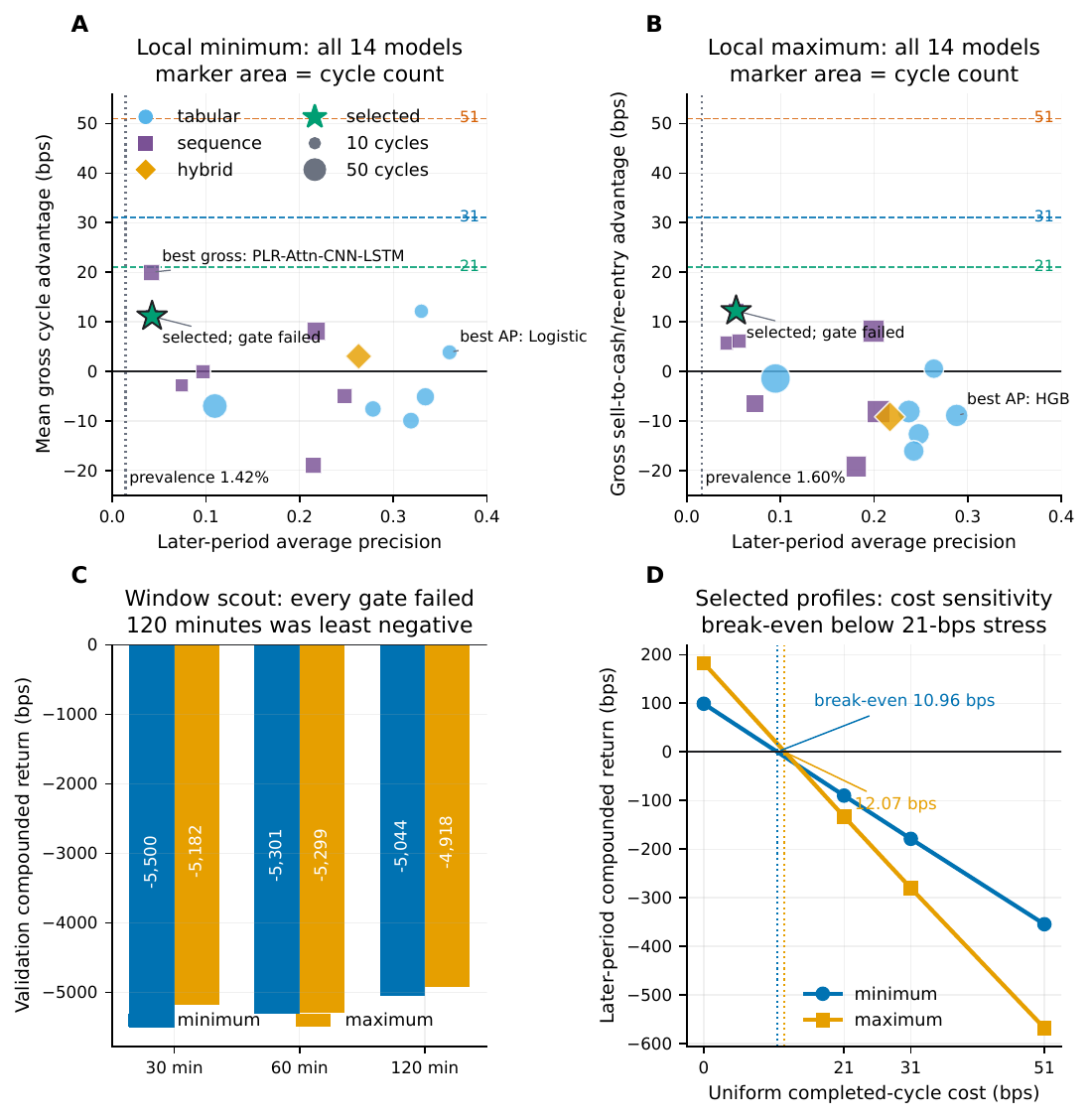}
  \caption{Complete local-extrema model and policy audit.  Panels A and B show all 14 minimum and all 14 maximum models on the July 1--12 model-specific evaluation: average precision is plotted against gross mean cycle advantage, marker area encodes completed-cycle support, model-family markers expose architecture class, and the validation-selected attention CNN--LSTM is identified separately from other evaluated models.  Event prevalence is 1.418\% for minima and 1.603\% for maxima; horizontal 21/31/51-bps lines show the uniform completed-cycle costs a gross mean would need to exceed.  Panel C reports the validation-only 30/60/120-minute window scout that froze the label radius before model comparison.  Panel D compounds the selected observed cycles at 0, 21, 31, and 51 bps and marks exact zero-return costs of 10.9616 bps (9 minimum cycles) and 12.0690 bps (15 maximum cycles).  These short, model-specific July slices were exposed to project-level research and do not provide independent model-selection or promotion evidence.}
  \label{fig:local-extrema}
\end{figure}

Figure~\ref{fig:local-extrema} expands the five-row summary into the complete 28-model local campaign.  It shows all fourteen architectures in each direction, so the selected attention CNN--LSTM and any high-AP or high-gross evaluated point can be judged against the full searched family.  No minimum or maximum model had a gross mean large enough to clear the 21-bps floor.  The exact uniform costs that set the selected observed compounded paths to zero were only 10.9616 and 12.0690 bps for minima and maxima, respectively; these are realized-sample break-even summaries of 9 and 15 cycles, not estimates of attainable future costs or edge.

The selected minimum and maximum models won 5/9 and 8/15 cycles.  Under the same independent-Bernoulli simplification, their descriptive Clopper--Pearson 95\% win-rate intervals are 21.2--86.3\% and 26.6--78.7\%.  Their arithmetic gross mean advantages were 11.11 and 12.21 bps/cycle, close to but slightly above the exact compounded break-even costs because arithmetic averaging ignores path compounding.  Both advantages were below the 21-bps stress and therefore negative before the primary 31-bps assumption.  Win count alone hid asymmetric target/stop geometry.

\clearpage

The paired daily adaptation sharpens the same point.  Minimum event AUC/AP/prevalence was 0.8742/0.1341/2.983\%; maximum was 0.8962/0.1158/2.131\%.  Yet six of seven completed cycles lost.  At 31 bps, portfolio return was -44.30\%, versus -41.20\% for buy-and-hold, an excess of -3.10 percentage points.  Lowering cost to 21 bps changed the strategy return only to -44.10\%, so fees alone do not explain the failure.  Across the seven realized cycles, the sole positive cycle returned +15.55\%, while the worst returned -29.56\%; terminal-open liquidation is included in the reported path.

\begin{figure}[htbp]
  \centering
  \includegraphics[width=\textwidth]{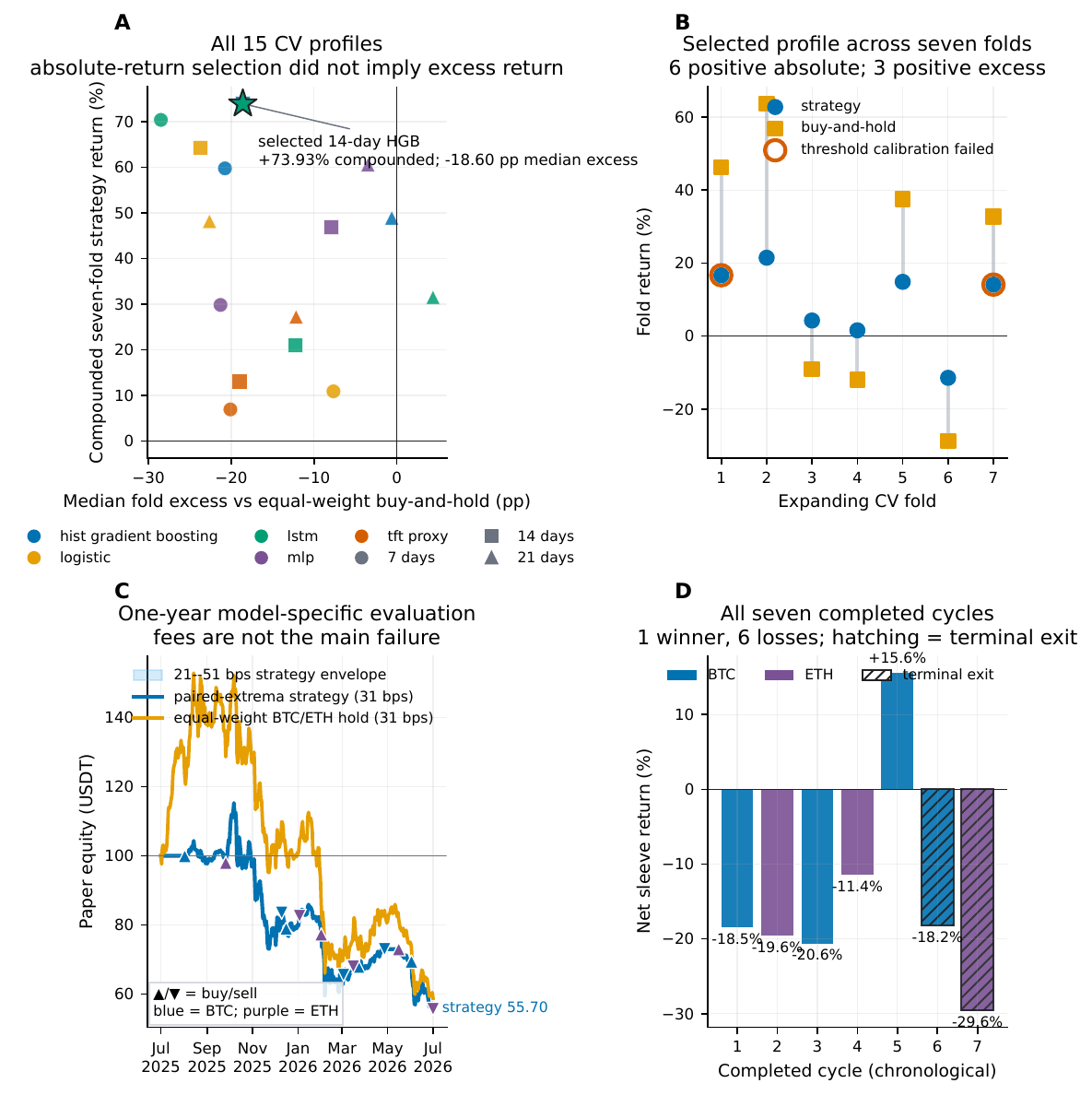}
  \caption{Gurgul-inspired OHLCV-only daily adaptation.  Panel A relates median buy-and-hold excess to compounded return for all 15 profiles over seven expanding folds; the absolute-return-selected 14-day HGB proxy had -18.60 percentage points median excess.  Panel B separates strategy return from benchmark return by fold.  Panel C traces 366 daily marks from July 2025 through June 2026 under a 31-bps round-trip convention applied as 15.5 bps multiplicatively per executed side; the 21--51-bps envelope and BTC/ETH hold benchmark are cost-matched.  Up/down triangles denote buy/sell and blue/purple denote BTC/ETH.  Panel D reports all seven completed cycles (six negative), distinguishing terminal exits.  From 100 USDT, terminal values are 55.70 for the strategy and 58.80 for hold.  This two-sleeve public-OHLCV proxy is not a reproduction of the source paper's multimodal models.}
  \label{fig:paired-daily}
\end{figure}
\clearpage

\subsection{Adaptive search and the positive rotation control}

Table~\ref{tab:search} makes the researcher's degrees of freedom visible.  These counts rule out treating a best validation row as an independent discovery.

\begin{table}[htbp]
  \centering
  \caption{Adaptive search ledger.  The prospective extension introduces no new choice.}
  \label{tab:search}
  \resizebox{\textwidth}{!}{\begin{tabular}{lll}
\toprule
Campaign & Search budget & Interpretation \\
\midrule
One4All repartitioned diagnostic & 3,996 profile combinations & Boundary-overlapped; narrative only \\
Mandatory-daily predecessor & 946 enumerated across 8 campaigns; earlier searches additional & April--June reused for cross-family freeze \\
Local-minimum campaign & 238 validation policy comparisons & Window, model, and threshold selected before July \\
Local-maximum campaign & 238 validation policy comparisons & Window, model, and threshold selected before July \\
Daily paired-extrema adaptation & 15 profiles x 7 folds; 7-threshold grids & Absolute-return selection; relative weakness visible \\
Slow rotation audit & 240 policies & Consumed historical selector; descriptive only \\
July prospective extension & 0 new choices & Loaded frozen model; retained all later outcomes \\
\bottomrule
\end{tabular}
}
\end{table}

The consumed rotation baseline is historically positive: +313.01\% under the 31-bps round-trip turnover convention over 45 scored months, compared with +78.69\% for equal-weight hold.  It is not promotion evidence.  Only 15 months were target-active, 8 of 21 traded months were positive, and November 2024 returned +160.422\% and supplied 67.48\% of total net log gain.  Removing that month lowers reconstructed terminal wealth from 413.01 to 158.59 USDT.  A predeclared two-asset cap reduced maximum drawdown from -48.38\% to -38.69\%, but also reduced full-path wealth to 350.22 USDT; its November return was +90.954\%, largest-month log-gain share was 51.61\%, and leave-one-month-out terminal wealth was 183.40 USDT.  Cap-minus-baseline log return was -0.164925, with a one-sided 90\% four-month moving-block lower bound of -0.738153.  Concentration, sparse support, consumed data, and mechanical universe selection leave both profiles at \texttt{NO\_TRADE}.

\begin{figure}[htbp]
  \centering
  \includegraphics[width=\textwidth]{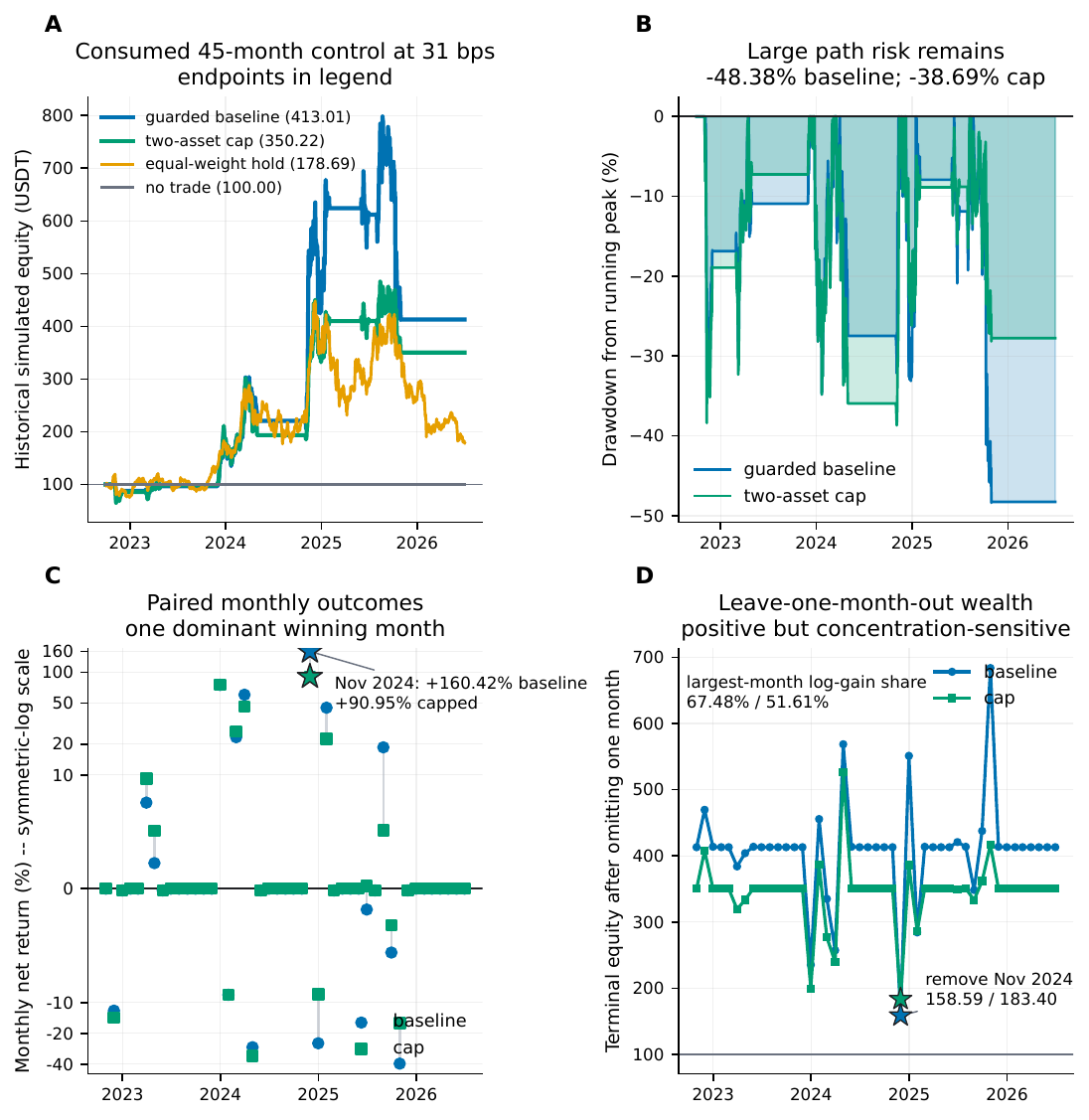}
  \caption{Consumed slow-rotation control.  The primary 31-bps round-trip label is charged as half-cost per unit of one-way turnover.  Panel A shows 1,370 common daily timestamps per series; terminal wealth is 413.01 USDT (baseline), 350.22 (two-asset cap), 178.69 (equal-weight hold), and 100 (no trade).  Panel B shows -48.38\%/-38.69\% maximum drawdowns for baseline/cap.  Panel C reports all 45 monthly returns for each profile, exposing November 2024 concentration (+160.422\%/+90.954\%).  Panel D omits each month in turn; removing November leaves 158.59/183.40 USDT.  All data are consumed; a twelve-month trailing selector compared 240 policies, and sparse activity, universe screening, and 67.48\%/51.61\% largest-month log-gain shares preclude deployment.}
  \label{fig:rotation-control}
\end{figure}
\clearpage

\section{Forensic Audit of the Nominal One4All Holdout}
\label{sec:forensic}

The previous manuscript called May 24--June 23 an unseen 30-day holdout.  That language was too strong for four reasons:
\begin{enumerate}[leftmargin=*]
  \item overlapping dates had already influenced the June 28 after-min outcome-head architecture;
  \item the runner built four-hour future labels before splitting and did not purge up to 48 five-minute rows at train--calibration or calibration--test boundaries;
  \item the simulator decided from the completed candle and filled at that same close, an optimistic convention without a market-on-close execution mechanism;
  \item the two claimed result directories are absent locally, while the earlier publication script silently substituted hard-coded summaries, loss counts, and synthetic fallback-generated equity paths.
\end{enumerate}

The period was frozen with respect to the final profile sweep, but it was not globally untouched and its boundary was overlapped.  This revision therefore calls it a \emph{repartitioned archival diagnostic}.  Its hard-coded result table, holdout-collapse figure, loss-asymmetry fallback, and synthetic fallback-generated rotation paths were removed.  The new asset pipeline fails if any required empirical input is missing.  A future quantitative reuse would require restored immutable artifacts and a rerun with complete-horizon masks, a four-hour label-horizon purge at both split boundaries, next-open entry, fixed seeds, and a predeclared promotion gate.

This correction also changes the role of the earlier 60-trade row.  It was a post-hoc coverage diagnostic among calibration-selected profiles, not the primary frozen profile.  It cannot serve as independent evidence that one more threshold would have solved or disproved the strategy.

\section{Interpretation}

\subsection{Event annotation is not policy value}

The results do not show that extrema are intrinsically unpredictable.  Some detectors rank completed extrema well.  They show that the tested annotation targets do not align with the value of the executed actions.  Minimum labels reward recognizing a bar near a completed local low; entry value depends on the next open and subsequent target/stop/timeout path.  Maximum labels reward peak annotation; an exit decision depends on the value of holding versus entering cash and later repurchasing.  The paired daily policy further depends on state, signal conflict, long delays, and terminal liquidation.  This is a predict-then-optimize failure, not merely an AUC failure.

\subsection{Coverage can destroy a sparse edge}

Mandatory daily selection guarantees activity but removes abstention.  The 19-day result shows what happens when the chosen action is negative on average.  Earlier sparse validation pockets cannot be compared directly because they used other periods and search budgets, but they motivate a future selective-risk design: predeclare a cash action, calibrate acceptance on past-only data, and report the full coverage--risk curve rather than one favorable threshold.

\subsection{Why the result does not contradict positive studies}

Bysik and \mbox{\'{S}lepaczuk} find selected profitable hourly BTC policies after a cost-aware filter \citep{bysik2026costs}.  The difference may reflect horizon, universe, features, model, assumed cost, turnover, and selection protocol.  Their finding prevents a universal ``crypto ML cannot work'' conclusion; the present evidence supports only the contextual statement that these intraday multi-pair extrema and mandatory-action policies did not establish positive value.

\section{Threats to Validity}

\paragraph{Internal validity.}
The archival One4All diagnostic has known four-hour boundary overlap and same-close entry; it is not used as confirmatory evidence.  OHLC bars obscure intrabar ordering.  All candle simulations omit spread, slippage, latency, queue priority, partial fills, outages, and market impact.  Source-exchange corrections or delistings may change reconstructed samples.

\paragraph{Statistical validity.}
Search budgets are large, many deep models use one seed, and support is small: 19 later-period daily cycles, 9/15 selected local-extrema cycles, and 7 paired daily cycles.  The final July selector also followed at least 946 enumerated cross-campaign comparisons on April--June, and its reporting cutoff was not preregistered.  Consecutive cycles may be serially dependent, so the binomial intervals are descriptive under an independence model rather than exact time-series uncertainty statements.  No formal family-wise multiple-testing correction is claimed, and the heterogeneous campaigns do not support a pooled significance test.  The rotation block bootstrap addresses one predeclared paired comparison only.

\paragraph{Construct validity.}
ROC AUC measures ranking, not calibration or economic utility.  Balanced sampling and weighted losses do not make scores calibrated probabilities.  The paper therefore uses ``score'' unless calibration was measured and reports AP/prevalence for rare labels.  Transaction-cost stress labels are assumptions rather than realized all-in costs, and their application is campaign-specific: flat per completed intraday cycle, multiplicative half-cost per paired-daily side, or half-cost per unit of rotation turnover.

\paragraph{External validity.}
Results are specific to Binance Spot, selected USDT pairs, particular historical regimes, long-only/full-cash policies, and candle features.  Mechanical universe selection can introduce survivorship and look-ahead bias.  The Gurgul-inspired campaign cannot make claims about the source paper's multimodal specification.

\paragraph{Reproducibility.}
The arXiv source bundle compiles the manuscript and contains audited tables and figures, but raw candles and large checkpoints are excluded.  A full companion release still needs a tagged immutable repository/archive, environment lock, public-download manifests, configuration hashes, and all cited result summaries.  Exact dates and symbols alone are insufficient to recreate an identical snapshot.

\section{AI Assistance, Skills, and Reproducibility}
\label{sec:ai-assistance}

\subsection{Disclosure and responsibility}

The model campaigns completed through July 14 and the unchanged frozen-model extension executed on July 20 were produced by scripted fixed-seed model runs and deterministic simulators.  On July 20, the author used the OpenAI Codex agent environment (GPT-5 family; the exact service build was not recorded in the workspace interface) to continue the literature review, commission separately tasked manuscript/method/reference critique passes, trace claims to local artifacts, orchestrate the frozen extension, write audit code, revise LaTeX, and prepare the source package.  This disclosure is motivated by scholarly transparency guidance and general AI risk-management principles \citep{zielinski2024chatbots,nist2024genai}; it does not claim full WAME conformance because a full prompt transcript and exact service build were not retained.

The author supplied the research objective, repository instructions, safety boundary, and final publication authority.  Agent outputs were not accepted as evidence by themselves: values were reconciled against JSON/CSV/model files, scripts were compiled or tested, figures were regenerated fail-closed, and separately tasked agents challenged chronology and citations.  This process found substantive errors in the earlier draft, including the overstated ``unseen'' label and silent plot fallbacks.  No conversational prompt transcript is claimed as an immutable research artifact; durable instructions, skills, execution plans, audit outputs, and experience documents are retained instead.

\subsection{Project-local skills used in this revision}

The repository's \path{.agents/skills/} directory codifies repeatable procedures.  Five skills were used in this July 20 revision; the shadow-supervisor skill exists for sanitized runtime review but was not used to generate, evaluate, or write the economic results.

\begin{table}[htbp]
  \centering
  \caption{AI-agent skills and their actual role in this manuscript revision.}
  \label{tab:skills}
  \begin{tabular}{>{\raggedright\arraybackslash}p{0.32\textwidth}>{\raggedright\arraybackslash}p{0.56\textwidth}}
    \toprule
    Skill & Use and boundary \\
    \midrule
    \mbox{\texttt{research-evidence-}}\linebreak\mbox{\texttt{synthesis}} & Located reports, summaries, trade files, and experience memory; separated positive controls, diagnostics, and rejected evidence. \\
    \mbox{\texttt{chronological-validation-}}\linebreak\mbox{\texttt{audit}} & Exposed prior date reuse, the missing four-hour purge, same-close entry, search-budget issues, and overstrong promotion language. \\
    \mbox{\texttt{experience-memory-}}\linebreak\mbox{\texttt{curator}} & Appended the frozen July 8--19 extension, exact artifacts, assumptions, negative result, and next decision. \\
    \mbox{\texttt{manuscript-}}\linebreak\mbox{\texttt{evidence-update}} & Rewrote claims from verified evidence, preserved \texttt{NO\_TRADE}, and removed unsupported figures and paths. \\
    \mbox{\texttt{arxiv-source-package}} & Checks metadata, referenced files, TeX compilation, and the minimal source ZIP; it cannot submit for the author. \\
    \mbox{\texttt{shadow-supervisor-review}} & Not used in the economic experiments or this evidence synthesis; reserved for sanitized, advisory, shadow-only runtime summaries. \\
    \bottomrule
  \end{tabular}
\end{table}

\subsection{Auditable files}

The core revision script is \path{latex-publication/scripts/audit_empirical_evidence.py}.  It has no numeric fallbacks and reconciles counts, compounded returns, costs, exact binomial intervals, the prospective model hash, rotation concentration, and missing narrative-only inputs.  Its source records include:
\begin{itemize}[leftmargin=*]
  \item \path{labs/results/all4one-1m-july-prospective-20260709/};
  \item \path{labs/results/all4one-1m-july-prospective-extension-20260720/};
  \item \path{labs/results/local-minima-model-comparison-20260714/};
  \item \path{labs/results/local-maxima-model-comparison-20260714/};
  \item \path{labs/results/gurgul-paired-daily-extrema-20260714-portable/};
  \item \path{labs/results/robust-regime-rotation/consumed-5y-20260705-v3/};
  \item \path{labs/results/paper-evidence-audit-20260720/summary.json};
  \item \path{docs/experiences/all4one-july-prospective-check-2026-07-09.md}.
\end{itemize}
The mutable \repositoryurl{} \citep{quantbot2026repo} is a discovery location, not yet an immutable archival DOI.  The paper therefore avoids claiming that the current public repository alone is sufficient for exact end-to-end reconstruction.

\section{Conclusion}

Within the tested, mostly exploratory protocols, event-ranking performance did not establish positive executable policy value.  The most defensible later-period result is the unchanged nineteen-cycle July campaign, which lost 6.72\% at 31 bps after extensive predecessor search.  The local-extrema and paired daily studies show that high ROC AUC can coexist with low average precision, weak gross edge, and negative stateful policy return.  The positive slow-rotation history remains a consumed, concentrated control rather than deployment evidence.

The AI-assisted revision made one contribution that is measurable in the record: it helped expose claims and figures that failed provenance or chronology checks.  Correcting those errors narrowed the conclusion but strengthened its credibility.  No evaluated policy passed its economic and evidence gates, so the operational conclusion remains \texttt{NO\_TRADE}.

Future work should pre-register a policy-conditioned target, a cash/abstain action, a point-in-time universe, full-horizon purges, next-executable-price timing, multiple seeds, all-in cost stresses, benchmark excess, drawdown and concentration gates, and a genuinely prospective evaluation window.  Additional threshold search on the consumed periods would not answer the question.

\clearpage
\appendix
\section{Complete Local-Extrema Model Diagnostics}

Tables~\ref{tab:all-minimum-models} and~\ref{tab:all-maximum-models} report every model shown in Figure~\ref{fig:local-extrema}.  AUC and AP summarize rare-event ranking; P/R is threshold-specific precision/recall; gross and net means are bps per completed cycle; and Return/$|\mathrm{DD}|$ gives compounded return and maximum-drawdown magnitude in percent at the primary 31-bps cost.  Only the attention CNN--LSTM row in each direction was selected by the validation protocol.  Every other July row is a post-hoc diagnostic and cannot replace the failed validation gate or support model promotion.

\begin{center}
  \captionsetup{hypcap=false}
  \captionof{table}{All 14 local-minimum models on the July 1--12 model-specific evaluation.  The selected row completed 9 cycles; all unselected rows are post-hoc diagnostics.  Returns are long-only next-open entries from cash under a 31-bps completed-cycle deduction.}
  \label{tab:all-minimum-models}
\end{center}
\begingroup
  \renewcommand{\theHtable}{minimum-internal-\arabic{table}}
  \scriptsize
  \setlength{\tabcolsep}{2.5pt}
  \renewcommand{\arraystretch}{1.08}
  \begin{tabularx}{\textwidth}{@{}>{\raggedright\arraybackslash}X l r r r r r r r r@{}}
\toprule
Model & Status & AUC & AP & P/R & Cycles & W/L & Gross mean & Net mean & Return / $|\mathrm{DD}|$ \\
\midrule
attention cnn lstm & selected & 0.729 & 0.043 & 0.10/0.04 & 9 & 5/4 & +11.1 & -19.9 & -1.79/2.07\% \\
logistic & diagnostic & 0.973 & 0.360 & 0.73/0.13 & 15 & 7/8 & +3.8 & -27.2 & -4.02/4.85\% \\
hist gradient boosting & diagnostic & 0.974 & 0.334 & 0.50/0.16 & 27 & 8/19 & -5.1 & -36.1 & -9.34/9.86\% \\
weighted linear svm & diagnostic & 0.973 & 0.330 & 0.68/0.10 & 13 & 7/6 & +12.1 & -18.9 & -2.44/3.29\% \\
random forest & diagnostic & 0.973 & 0.319 & 0.53/0.14 & 21 & 6/15 & -9.9 & -40.9 & -8.28/8.81\% \\
mlp & diagnostic & 0.968 & 0.278 & 0.41/0.09 & 20 & 7/13 & -7.6 & -38.6 & -7.46/8.32\% \\
gru feature boosting & diagnostic & 0.958 & 0.263 & 0.43/0.13 & 27 & 10/17 & +3.0 & -28.0 & -7.31/8.17\% \\
lstm & diagnostic & 0.959 & 0.248 & 0.45/0.07 & 14 & 4/10 & -4.9 & -35.9 & -4.93/5.75\% \\
gru & diagnostic & 0.958 & 0.217 & 0.35/0.08 & 26 & 12/14 & +8.2 & -22.8 & -5.80/6.83\% \\
cnn gru & diagnostic & 0.961 & 0.215 & 0.36/0.06 & 17 & 5/12 & -19.0 & -50.0 & -8.19/8.51\% \\
gaussian causal bayes & diagnostic & 0.921 & 0.109 & 0.14/0.27 & 57 & 18/39 & -7.0 & -38.0 & -19.56/19.56\% \\
cnn & diagnostic & 0.880 & 0.097 & 0.24/0.04 & 11 & 5/6 & +0.0 & -31.0 & -3.37/3.93\% \\
transformer & diagnostic & 0.835 & 0.074 & 0.14/0.03 & 7 & 3/4 & -2.9 & -33.9 & -2.36/2.41\% \\
plr attention cnn lstm & diagnostic & 0.627 & 0.042 & 0.09/0.03 & 18 & 11/7 & +20.0 & -11.0 & -1.98/2.27\% \\
\bottomrule
\end{tabularx}

  \addtocounter{table}{-1}
\endgroup

\begin{center}
  \captionsetup{hypcap=false}
  \captionof{table}{All 14 local-maximum models on the July 1--12 model-specific evaluation.  The selected row completed 15 cycles; all unselected rows are post-hoc diagnostics.  Return is the advantage of selling an already-held Spot position to cash and later repurchasing under a 31-bps completed-cycle deduction; it is neither short P\&L nor standalone cash-entry return.}
  \label{tab:all-maximum-models}
\end{center}
\begingroup
  \renewcommand{\theHtable}{maximum-internal-\arabic{table}}
  \scriptsize
  \setlength{\tabcolsep}{2.5pt}
  \renewcommand{\arraystretch}{1.08}
  \begin{tabularx}{\textwidth}{@{}>{\raggedright\arraybackslash}X l r r r r r r r r@{}}
\toprule
Model & Status & AUC & AP & P/R & Cycles & W/L & Gross mean & Net mean & Return / $|\mathrm{DD}|$ \\
\midrule
attention cnn lstm & selected & 0.750 & 0.053 & 0.13/0.05 & 15 & 8/7 & +12.2 & -18.8 & -2.80/2.91\% \\
hist gradient boosting & diagnostic & 0.969 & 0.288 & 0.37/0.17 & 45 & 14/31 & -8.9 & -39.9 & -16.50/16.50\% \\
mlp & diagnostic & 0.965 & 0.264 & 0.37/0.12 & 32 & 11/21 & +0.5 & -30.5 & -9.33/9.60\% \\
logistic & diagnostic & 0.968 & 0.248 & 0.28/0.13 & 39 & 10/29 & -12.7 & -43.7 & -15.73/15.73\% \\
weighted linear svm & diagnostic & 0.967 & 0.242 & 0.29/0.12 & 36 & 8/28 & -16.1 & -47.1 & -15.65/15.65\% \\
random forest & diagnostic & 0.963 & 0.237 & 0.29/0.16 & 47 & 13/34 & -8.1 & -39.1 & -16.86/16.86\% \\
gru feature boosting & diagnostic & 0.946 & 0.217 & 0.28/0.13 & 40 & 12/28 & -9.2 & -40.2 & -14.91/14.91\% \\
cnn gru & diagnostic & 0.951 & 0.205 & 0.22/0.12 & 43 & 13/30 & -8.1 & -39.1 & -15.53/15.53\% \\
gru & diagnostic & 0.941 & 0.200 & 0.28/0.13 & 41 & 19/22 & +8.1 & -22.9 & -9.02/9.06\% \\
lstm & diagnostic & 0.939 & 0.181 & 0.26/0.10 & 36 & 6/30 & -19.2 & -50.2 & -16.59/16.80\% \\
gaussian causal bayes & diagnostic & 0.901 & 0.095 & 0.12/0.36 & 84 & 30/54 & -1.4 & -32.4 & -23.97/23.97\% \\
cnn & diagnostic & 0.832 & 0.073 & 0.13/0.07 & 22 & 8/14 & -6.5 & -37.5 & -7.96/7.96\% \\
transformer & diagnostic & 0.793 & 0.056 & 0.09/0.03 & 12 & 5/7 & +6.1 & -24.9 & -2.97/2.97\% \\
plr attention cnn lstm & diagnostic & 0.616 & 0.043 & 0.18/0.04 & 11 & 5/6 & +5.7 & -25.3 & -2.77/2.77\% \\
\bottomrule
\end{tabularx}

  \addtocounter{table}{-1}
\endgroup
\clearpage

\bibliographystyle{plainnat}
\bibliography{references}

\end{document}